\documentclass[supp]{new_tlp}
\usepackage{supp}
\usepackage{aopmath}

\usepackage{mathptmx}
\usepackage{times}
\usepackage{helvet}
\usepackage{courier}
\usepackage{graphicx}
\usepackage{caption}
\usepackage{subcaption}

\newcommand{\lpnot}{\mbox{not}\;\,}

\newcommand{\hif}{\leftarrow}

\newcommand{\aspindent}{\hspace*{.5in}}

\title[Towards an ASP-Based Architecture for Autonomous UAVs in Dynamic Environments]{Towards an ASP-Based Architecture for Autonomous UAVs in Dynamic Environments (Extended Abstract)}
\author[M.~Balduccini, W.~Regli and D.~Nguyen]{Marcello Balduccini, William C.~Regli, Duc N.~Nguyen\\
Applied Informatics Group\\
Drexel University\\
Philadelphia, PA, USA
}

\begin{document}

\maketitle
\begin{abstract}
\begin{quote}

Traditional AI reasoning techniques have been used successfully in many domains, including logistics, scheduling and game playing.  This paper is part of a project aimed at investigating how such techniques can be extended to coordinate teams of unmanned aerial vehicles (UAVs) in dynamic environments.  Specifically challenging are real-world environments where UAVs and other network-enabled devices must communicate to coordinate -- and communication actions are neither reliable nor free.  Such network-centric environments are common in military, public safety and commercial applications, yet most research (even multi-agent planning) usually takes communications among distributed agents as a given. We address this challenge by developing an agent architecture and reasoning algorithms based on Answer Set Programming (ASP). Although ASP has been used successfully in a number of applications,  to the best of our knowledge this is the first practical application of a complete ASP-based agent architecture. It is also the first practical application of ASP involving a combination of centralized reasoning, decentralized reasoning, execution monitoring, and reasoning about network communications.

\noindent
\begin{center}
\emph{To appear in Theory and Practice of Logic Programming (TPLP).}
\end{center}

\end{quote}
\end{abstract}

\section{Introduction}
Unmanned Aerial Vehicles (UAVs) promise to revolutionize the way in which we use our airspace.  From talk of automating the navigation for major shipping companies to the use of small helicopters as "deliverymen" that drop your packages at the door, it is clear that our airspaces will become increasingly crowded in the near future.  This increased utilization and congestion has created the need for new and different methods of coordinating assets using the airspace.  Currently, airspace management is the job for mostly human controllers. As the number of entities using the airspace vastly increases---many of which are autonomous---the need for improved autonomy techniques becomes evident. 

The challenge in an environment full of UAVs is that the world is highly dynamic and the communications environment is uncertain, making coordination difficult. Communicative actions in such setting are neither reliable nor free. 

The work discussed here is in the context of the development of a novel application of network-aware reasoning and of an intelligent mission-aware network layer to the problem of UAV coordination. Typically, AI reasoning techniques do not consider realistic network models, nor does the network layer reason dynamically about the needs of the mission plan. With network-aware reasoning, a reasoner (either centralized or decentralized) factors in the communications network and its conditions.

In this paper we provide a general overview of the approach, and then focus on the aspect of network-aware reasoning. We address this challenge by developing an agent architecture and reasoning algorithms based on Answer Set Programming (ASP, \cite{gl91,mt99,bar03}). ASP has been chosen for this task because it enables high flexibility of representation, both of knowledge and of reasoning tasks. Although ASP has been used successfully in a number of applications, to the best of our knowledge this is the first practical application of a complete ASP-based agent architecture. It is also the first practical application of ASP involving a combination of centralized reasoning, decentralized reasoning, execution monitoring, and reasoning about network communications.

The next section describes relevant systems and reasoning techniques, and is followed by a motivating scenario that applies to UAV coordination. The Technical Approach section describes network-aware reasoning and demonstrates the level of sophistication of the behavior exhibited by the UAVs using an example problem instance.
Finally, we draw conclusions and discuss future work.

\section{Related Work}\label{sec:background}
Incorporating network properties into planning and decision-making has been investigated in~\cite{usbeckCR12}. The authors' results indicate that plan execution effectiveness and performance is increased with the increased network-awareness during the planning phase. The UAV coordination approach in this current work combines network-awareness during the reasoning processes with a plan-aware network layer. 

The problem of mission planning for UAVs under communication constraints has been addressed in~\cite{kpj13}, where an ad-hoc task allocation process is employed to engage under-utilized UAVs as communication relays. In our work, we do not separate planning from the engagement of under-utilized UAVs, and do not rely on ad-hoc, hard-wired behaviors. Our approach gives the planner more flexibility and finer-grained control of the actions that occur in the plans, and allows for the emergence of sophisticated behaviors without the need to pre-specify them.

The architecture adopted in this work is an evolution of \cite{bg08}, which can be viewed as an instantiation of the BDI agent model \cite{rg91,woo00}. Here, the architecture has been extended to include a centralized mission planning phase, and to reason about other agents' behavior. Recent related work on logical theories of intentions \cite{bgb14} can be further integrated into our approach to allow for a more systematic hierarchical characterization of  actions, which is likely to increase performance.   

Traditionally, AI planning techniques have been used (to great success) to perform  multi-agent teaming, and UAV coordination. Multi-agent teamwork decision frameworks such as the ones described in \cite{pt02} may factor communication costs into the decision-making. However, the agents do not actively reason about other agent's observed behavior, nor about the communication process. Moreover, policies are used as opposed to online reasoning about models of domains and of agent behavior. 

The reasoning techniques used in the present work have already been successfully applied to domains ranging from complex cyber-physical systems to workforce scheduling. To the best of our knowledge, however, they have never been applied to domains combining realistic communications and multiple agents.   

Finally, high-fidelity multi-agent simulators (e.g., AgentFly~\cite{sislak2012agentfly}) do not account for network dynamism nor provide a realistic network model. For this reason, we base our simulator on the Common Open Research Emulator (CORE)~\cite{ahrenholz10core}. CORE provides network models in which communications are neither reliable nor free.

\section{Motivating Scenario}\label{sec:scenario}
To motivate the need for network-aware reasoning and mission-aware networking, consider a simple UAV coordination problem, depicted in Figure~\ref{fig:inst3}, in which two UAVs are tasked with taking pictures of a set of three targets, and with relaying the information to a home base.

Fixed relay access points extend the communications range of the home base. The UAVs can share images of the targets with each other and with the relays when they are within radio range. The simplest solution to this problem consists in entirely disregarding the networking component of the scenario, and generating a mission plan in which each UAV flies to a different set of targets, takes pictures of them, and flies back to the home base, where the pictures are transferred. This solution, however, is not satisfactory. First of all, it is inefficient, because it requires that the UAVs fly all the way back to the home base before the images can be used. The time it takes for the UAVs to fly back may easily render the images too outdated to be useful. 
Secondly, disregarding the  network during the reasoning process may lead to mission failure --- especially in the case of unexpected events, such as enemy forces blocking transit to and from the home base after a UAV has reached a target. Even if the UAVs are capable of autonomous behavior, they will not be able to complete the mission unless they take advantage of the network.

Another common solution consists of acknowledging the availability of the network, and assuming that the network is constantly available throughout plan execution. A corresponding mission plan would instruct each UAV to fly to a different set of targets, and take pictures of them, while the network relays the data back to the home base. This solution is \emph{optimistic} in that it assumes that the radio range is sufficient to reach the area where the targets are located, and that the relays will work correctly throughout the execution of the mission plan.

This optimistic solution is more efficient than the previous one, since the pictures are received by the home base soon after they are taken. Under realistic conditions, however, the strong assumptions it relies upon may easily lead to mission failure---for example, if the radio range does not reach the area where the targets are located. 

 In this work, the reasoning processes take into account not only the presence of the network, but also its configuration and characteristics, taking advantage of available resources whenever possible. The mission planner is given information about the radio range of the relays and determines, for example, that the targets are out of range. A possible mission plan constructed by this information into account consists in having one UAV fly to the targets and take pictures, while the other UAV remains in a position to act as a network bridge between the relays and the UAV that is taking pictures. This solution is as efficient as the optimistic solution presented earlier, but is more robust, because it does not rely on the same strong assumptions.  

Conversely, when given a mission plan, an intelligent network middleware service capable of sensing conditions and modifying network parameters (e.g., modify network routes, limit bandwidth to certain applications, and prioritize network traffic) is able to adapt the network to provide optimal communications needed during plan execution. A relay or UAV running such a middleware is able to interrupt or limit bandwidth given to other applications to allow the other UAV to transfer images and information toward home base. Without this traffic prioritization, network capacity could be reached prohibiting image  transfer.

\section{Technical Approach}
In this section, we formulate the problem in more details, discuss the design of the agent architecture and of the reasoning modules, and demonstrate the sophistication of the resulting behavior of the agents in two scenarios. We assume familiarity with ASP, and refer the reader to \cite{gl91,ns00,bar03} for an introduction on the topic.

\subsection{Problem Formulation}

A problem instance for coordinating UAVs to observe targets and deliver information (e.g., images) to a home base is defined by a set of UAVs, $u_1, u_2, \ldots$, a set of targets, $t_1, t_2, \ldots$, a (possibly empty) set of fixed radio relays, $r_1, r_2, \ldots$, and a home base. The UAVs, the relays, and the home base are called radio nodes (or network nodes). Two nodes are in  radio contact if they are within a distance $\rho$ from each other, called radio range\footnote{For simplicity, we assume that all the radio nodes use comparable network devices, and that thus $\rho$ is unique throughout the environment.}, or if they can relay information to each other through intermediary radio nodes that are themselves within radio range. The UAVs are expected to travel from the home base to the targets to take pictures of the targets and deliver them to the home base. A UAV will automatically take a picture when it reaches a target. If a UAV is within radio range of a radio node, the pictures are automatically shared.
From the UAVs' perspective, the environment is only partially observable. Features of the domain that are observable to a UAV $u$ are (1) which radio nodes $u$ can and cannot communicate with by means of the network, and (2) the position of any UAV that is near $u$. 

The goal is to have the UAVs take a picture of each of the targets so that (1) the task is accomplished as quickly as possible, and (2) the total ``staleness'' of the pictures is as small as possible. Staleness is defined as the time elapsed from the moment a picture is taken, to the moment it is received by the home base. While the UAVs carry on their tasks, the relays are expected to actively prioritize traffic over the network in order to ensure mission success and further reduce staleness.

\subsection{Agent Architecture}

The architecture used in this project follows the BDI agent model \cite{rg91,woo00},
which provides a good foundation because of its logical underpinning, clear structure and flexibility. In particular, we build upon ASP-based instances of this model \cite{bg00,bg08} because they employ directly-executable logical languages  featuring good computational properties while at the same time ensuring elaboration tolerance \cite{mcc98} and elegant handling of incomplete information, non-monotonicity, and dynamic domains.

\begin{figure}[htbp]
\hspace*{-.5in}
\begin{minipage}[c]{.55\linewidth}
  \begin{subfigure}[t]{\linewidth}
    \centering
        \includegraphics[width=1.1\columnwidth]{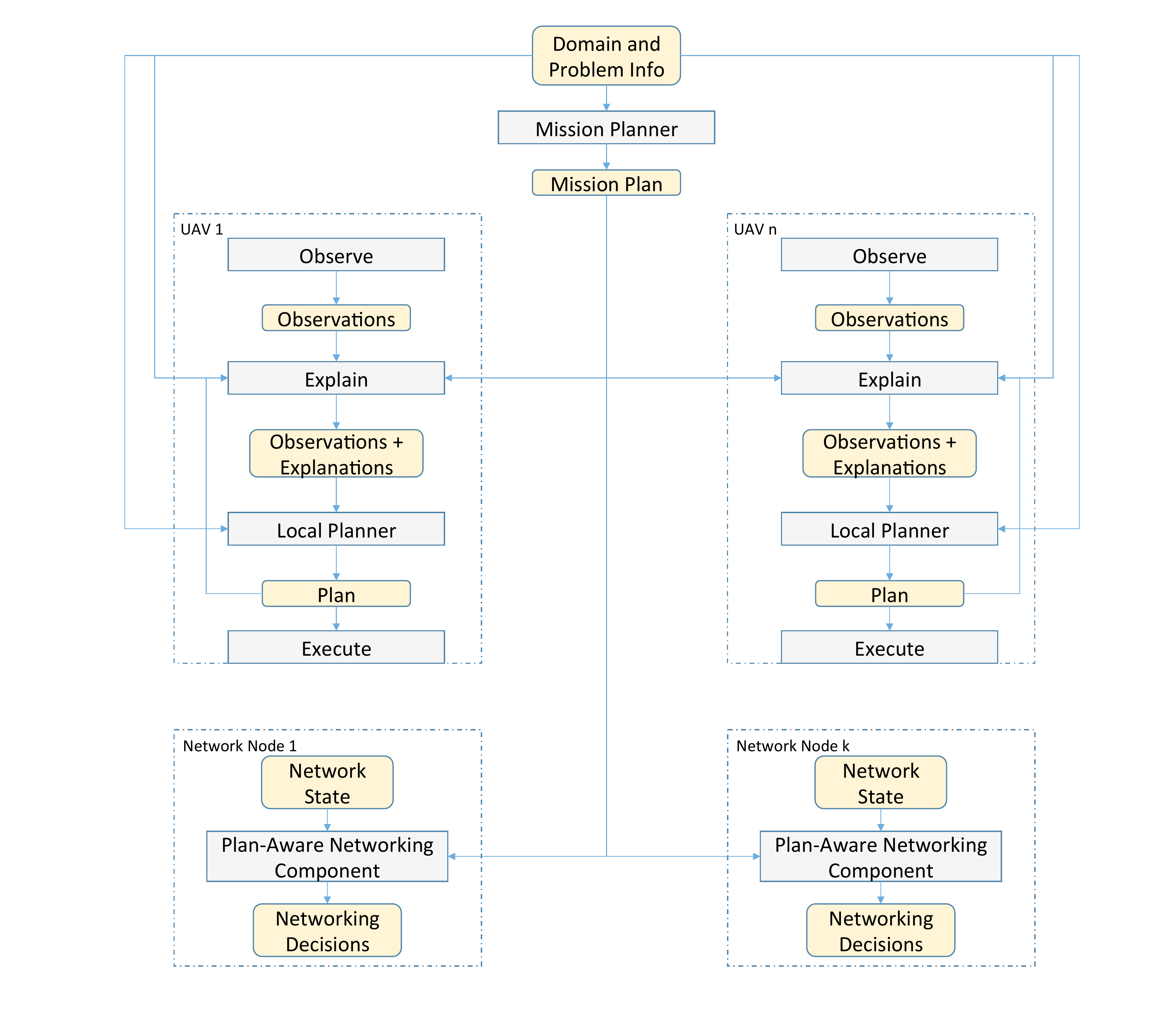}
  \end{subfigure}\\
\end{minipage}
\quad\quad
\begin{minipage}[c]{.4\linewidth}
\begin{subfigure}[t]{\linewidth}
\begin{center}
\begin{tabbing}
iiii\=iiii\=iiii\=iiii\=iiii\=iiii\=iiii\=iiii\=iiii\kill
\textbf{Input:} \>\>\> $M$: mission plan;\\
\>\>\>$G$: mission goal;\\
\textbf{Vars:}\>\>\>$H$ : history; \\
\>\>\>$P$: current plan; \\
\\
\>$P := M$; \\
\>$H := \mathrm{New\_Observations}()$; \\
\>\textbf{while} $\neg \mathrm{Goal\_Achieved}(H,G)$ \textbf{do}\\
\>\>\textbf{if} $\mathrm{Unexpected\_Observations}(H)$ \textbf{then}\\
\>\>\>$H := \mathrm{Explain\_Observations}(H)$; \\
\>\>\>$P := \mathrm{Compute\_Plan}(G,H,P)$; \\
\>\>\textbf{end if}\\
\>\>$A := \mathrm{Next\_Action}(P)$; \\
\>\>$P := \mathrm{Tail}(P)$; \\
\>\>$\mathrm{Execute}(A)$; \\
\>\>$H := \mathrm{Record\_Execution}(H,A)$; \\
\>\>$H := H \cup \,\mathrm{New\_Observations}()$; \\
\>\textbf{loop}\\
\end{tabbing}
\end{center}
\end{subfigure}
\end{minipage}
\caption{(a) Information flow \emph{(left)}; (b) Agent Control Loop \emph{(right)}.}
\label{fig:arch}
\label{fig:control-loop}
\end{figure}

A sketch of the information flow throughout the system is shown in Figure~\ref{fig:arch}a.\footnote{The tasks in the various boxes are executed only when necessary.} Initially, a centralized \emph{mission planner} is given a description of the domain and of the problem instance, and finds a plan that uses the available UAVs to achieve the  goal.

Next, each UAV receives the plan and begins executing it individually.
As plan execution unfolds, the communication state changes, potentially affecting network connectivity. For example, the UAVs may move in and out of range of each other and of the other network nodes. Unexpected events, such as relays failing or temporarily becoming disconnected, may also affect network connectivity. When that happens, each UAV reasons in a decentralized, autonomous fashion to overcome the issues. As mentioned earlier, the key to taking into account, and hopefully compensating for, any unexpected circumstances is to actively employ, in the reasoning processes, realistic and up-to-date information about the communications state.

The control loop used by each UAV is shown in Figure~\ref{fig:control-loop}b. In line with \cite{gl91,mt99,bar03}, the loop and the I/O functions are implemented procedurally, while the reasoning functions ($Goal\_Achieved$, $Unexpected\_Observations$, $Explain\_Observations$, $Compute\_Plan$) are implemented in ASP. The loop takes in input the mission goal and the mission plan, which potentially includes courses of actions for multiple UAVs. Functions $\mathrm{New\_Observations}$, $\mathrm{Next\_Action}$, $\mathrm{Tail}$, $\mathrm{Execute}$, $\mathrm{Record\_Execution}$ perform basic manipulations of data structures, and interface the agent with the  execution and perception layers. Functions $Next\_Action$ and $\mathrm{Tail}$ are assumed to be capable of identifying the portions of the mission plan that are relevant to the UAV executing the loop. The remaining functions occurring in the control loop implement the reasoning tasks. Central to the architecture is the maintenance of a history of past observations and actions executed by the agent. Such history is stored in variable $H$ and updated by the agent when it gathers observations about its environment and when it performs actions. It is important to note that variable $H$is local to the specific agent executing the loop, rather than shared among the UAVs (which would be highly unrealistic in a communication-constrained environment). Thus,  different agents will develop differing views of the history of the environment as execution unfolds. At a minimum, the difference will be due to the fact that agents cannot observe each other's actions directly, but only their consequences, and even those are affected by the partial observability of the environment.

Details on the control loop can be found in \cite{bg08}. With respect to that version of the loop, the control loop used in the present work does not allow for the selection of a new goal at run-time, but it extends the earlier control loop with the ability to deal with, and reason about, an externally-provided, multi-agent plan, and to reason about other agents' behavior. We do not expect run-time selection of goals to be difficult to embed in the control loop presented here, but doing so is out of the scope of the current phase of the project.   

\subsection{Network-Aware Reasoning}
The major reasoning tasks (centralized mission planning, as well as anomaly detection, explanation and planning within each agent) are reduced to finding models of answer-set based
formalizations of the corresponding problems.
Central to all the reasoning tasks is the ability to represent the evolution of the environment over time. Such evolution is conceptualized into a \emph{transition diagram} \cite{gl93}, a graph whose nodes correspond to states of the environment, and whose arcs describe state transitions due to the execution of actions. Let $\mathcal{F}$ be a collection of \emph{fluents}, expressions representing relevant properties of the domain that may change over time, and let $\mathcal{A}$ be a collection of \textit{actions}. A \emph{fluent literal} $l$ is a fluent $f \in \mathcal{F}$ or its negation $\neg f$. A \emph{state }$\sigma$ is a complete and consistent set of fluent literals. 

The transition diagram is formalized in ASP by rules describing the direct effects of actions, their executability conditions, and their indirect effects (also called state constraints). The succession of moments in the evolution of the environment is characterized by discrete \emph{steps}, associated with non-negative integers. The fact that a certain  fluent $f$ is true at a step $s$ is encoded by an atom $h(f,s)$. If $f$ is false, this is expressed by $\neg h(f,s)$. The occurrence of an action $a \in \mathcal{A}$ at step $s$ is represented as $o(a,s)$.

The history of the environment is formalized in ASP by two types of statements: $obs(f,true,s)$ states that $f$ was observed to be true at step $s$ (respectively, $obs(f,false,s)$ states that $f$ was false); $hpd(a,s)$ states that $a$ was observed to occur at $s$. Because in the this paper other agents' actions are not observable, the latter expression is used  only to record an agent's own actions.

Objects in the UAV domain discussed in this paper are the home base, a set of fixed relays, a set of UAVs, a set of targets, and a set of waypoints. The waypoints are used to simplify the path-planning task, which we do not consider in the present work. The locations that the UAVs can occupy and travel to are the home base, the waypoints, and the locations of targets and fixed relays. The current location, $l$, of UAV $u$ is represented by a fluent $at(u,l)$. For each location, the collection of its neighbors is defined by relation $next(l,l')$. UAV motion is restricted to occur only from a location to a neighboring one. The direct effect of action $move(u,l)$, intuitively stating that UAV $u$ moves to location $l$, and its executability condition are described by the following rules:
\[
\begin{array}{l}
h(at(U,L2),S+1) \hif \\
\aspindent o(move(U,L2),S), \\
\aspindent h(at(U,L1),S), \\
\aspindent next(L1,L2). \\
\ \\
\hif o(move(U,L2),S), \\
\hspace*{.18in}h(at(U,L1),S),\\
\hspace*{.18in}\lpnot next(L1,L2).
\end{array}
\]
The fact that two radio nodes are in radio contact is encoded by fluent $in\_contact(r_1,r_2)$.
The next two rules provide a recursive definition of the fluent, represented by means of state constraints:
\[
\begin{array}{l}
h(in\_contact(R1,R2),S) \hif \\
\aspindent R1 \neq R2, \ 
\neg h(down(R1),S), \ \neg h(down(R2),S), \\
\aspindent h(at(R1,L1),S), \ h(at(R2,L2),S), \ 
range(Rg), \ dist2(L1,L2,D), \ D \leq Rg^2. \\
\\
h(in\_contact(R1,R3),S) \hif \\
\aspindent R1 \neq R2,\ R2 \neq R3,\ R1 \neq R3, \ 
\neg h(down(R1),S),\ \neg h(down(R2),S), \\
\aspindent h(at(R1,L1),S),\ h(at(R2,L2),S), \\
\aspindent range(Rg),\ dist2(L1,L2,D),\ D \leq Rg^2,\\
\aspindent h(in\_contact(R2,R3),S).
\end{array}
\]
The first rule defines the base case of two radio nodes that are directly in range of each other. Relation $dist2(l_1,l_2,d)$ calculates the square  of the distance between two locations. Fluent $down(r)$ holds if radio $r$ is known to be out-of-order, and a suitable axiom (not shown) defines the closed-world assumption on it. In the formalization, $in\_contact(R1,R2)$ is a \emph{defined positive fluent}, i.e., a fluent whose truth value, in each state, is completely defined by the current value of other fluents, and is not subject to inertia. The formalization of $in\_contact(R1,R2)$ is thus completed by a rule capturing the closed-world assumption on it:
\[
\begin{array}{l}
\neg h(in\_contact(R1,R2),S) \hif \\
\aspindent R1 \neq R2,\\
\aspindent \lpnot h(in\_contact(R1,R2),S).
\end{array}
\]
Functions $\mathrm{Goal\_Achieved}$ and $\mathrm{Unexpected\_Observations}$, in Figure~\ref{fig:control-loop}a, respectively check if the goal has been achieved, and whether the history observed by the agent contains any unexpected observations. Following the definitions from \cite{gb02}, observations are unexpected if they contradict the agent's expectations about the corresponding state of the environment. This definition is captured by the \emph{reality-check axiom}, consisting of the constraints:
\[
\begin{array}{l}
\hif obs(F,true,S),\ \neg h(F,S). \\
\hif obs(F,false,S),\ h(F,S).
\end{array}
\]
Function $\mathrm{Explain\_Observations}$ uses a diagnostic process along the lines of \cite{gb02} to identify a set of exogenous actions (actions beyond the control of the agent that may occur unobserved), whose occurrence explains the observations. To deal with the complexities of reasoning in a dynamic, multi-agent domain, the present work extends the previous results on diagnosis by considering multiple types of exogenous actions, and preferences on the resulting explanations. The simplest type of exogenous action is $break(r)$, which occurs when radio node $r$ breaks. This action causes fluent $down(r)$ to become true. Actions of this kind may be used to explain unexpected observations about the lack of radio contact. However, the agent must also be able to cope with the limited observability of the position and motion of the other agents. This is accomplished by encoding commonsensical statements (encoding omitted) about the behavior of other agents, and about the factors that may affect it. The first such statement says that \emph{a UAV will normally perform
the mission plan, and will stop
performing actions when its portion of the mission plan is complete.} Notice that a mission plan is simply a sequence of actions. There is no need to include pre-conditions for the execution of the actions it contains, because those can be easily identified by each agent, at execution time, from the formalization of the domain.

The agent is allowed to hypothesize that \emph{a UAV  may have stopped executing the mission plan} (for example, if the UAV malfunctions or is destroyed).  \emph{Normally, the reasoning agent will expect a UAV that aborts execution to remain in its latest location.} In certain circumstances, however, a UAV may need to deviate completely from the mission plan. To accommodate for this situation, the agent may hypothesize that \emph{a UAV began behaving in an unpredictable way (from the agent's point of view) after aborting plan execution}. The following choice rule allows an agent to consider all of the possible explanations:
\[
\begin{array}{l}
\{\ hpd(break(R),S) , hpd(aborted(U,S)), hpd(unpredictable(U,S)) \ \}.
\end{array}
\]
A constraint ensures that unpredictable behavior can be considered only if a UAV is believed to have aborted the plan. If that happens,  the following choice rule is used to consider all  possible courses of actions from the moment the UAV became unpredictable to the current time step.
\[
\begin{array}{l}
\{ hpd(move(U,L),S') : S' \geq S : S' < currstep \} \hif hpd(unpredictable(U,S)).
\end{array}
\]In practice, such a thought process is important to enable coordination with other UAVs when communications between them are impossible, and to determine the side-effects of the inferred courses of actions and potentially take advantage of them (e.g., ``the UAV must have flown by target $t_3$. Hence, it is no longer necessary to take a picture of $t_3$''). A \emph{minimize} statement ensures that only cardinality-minimal diagnoses are found:
\[
\#minimize[ hpd(break(R),S), hpd(aborted(U,S)), hpd(unpredictable(U,S)) ].
\]

An additional effect of this statement is that the reasoning agent will prefer simpler explanations, which assume that a UAV aborted the execution of the mission plan and stopped, over those hypothesizing that the UAV engaged in an unpredictable course of actions.

Function $\mathrm{Compute\_Plan}$, as well as the mission planner, compute a new plan using a rather traditional approach, which relies on a choice rule for generation of candidate sequences of actions, constraints to ensure the goal is achieved, and \emph{minimize} statements to ensure  optimality of the plan with respect to the given metrics. 

Next, we outline a scenario demonstrating the features of our approach, including the ability to work around unexpected problems autonomously.

\begin{figure}[htbp]
  \centering
  \begin{subfigure}[t]{0.30\linewidth}
    \centering
    \includegraphics[width=\textwidth]{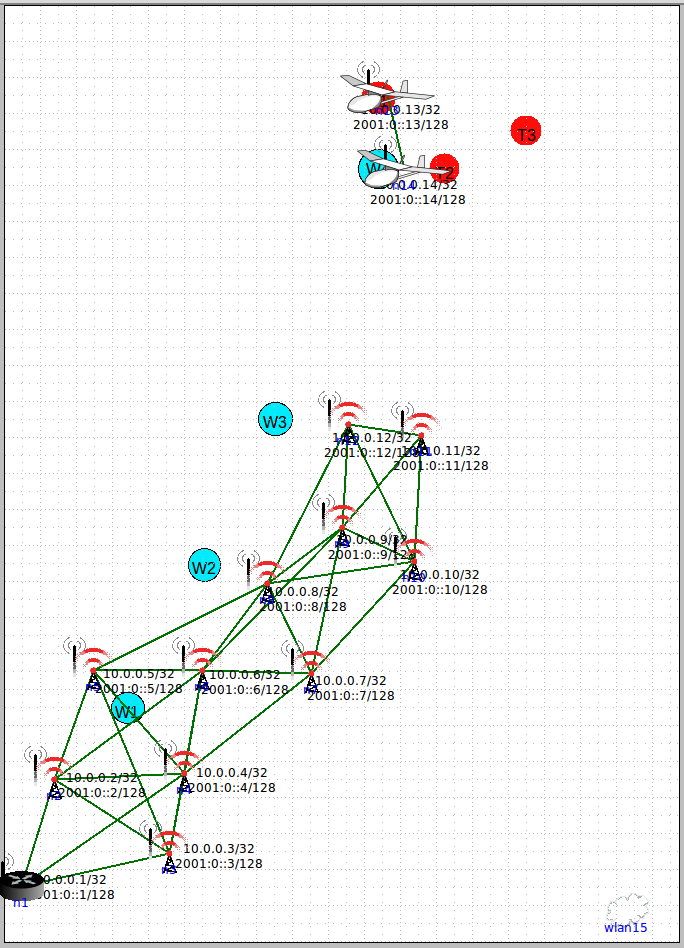}
    \subcaption{Step 5: $u_1$ transmitting to $u_2$.}\label{subfig:relays}
  \end{subfigure}
  \begin{subfigure}[t]{0.30\linewidth}
    \centering
    \includegraphics[width=\textwidth]{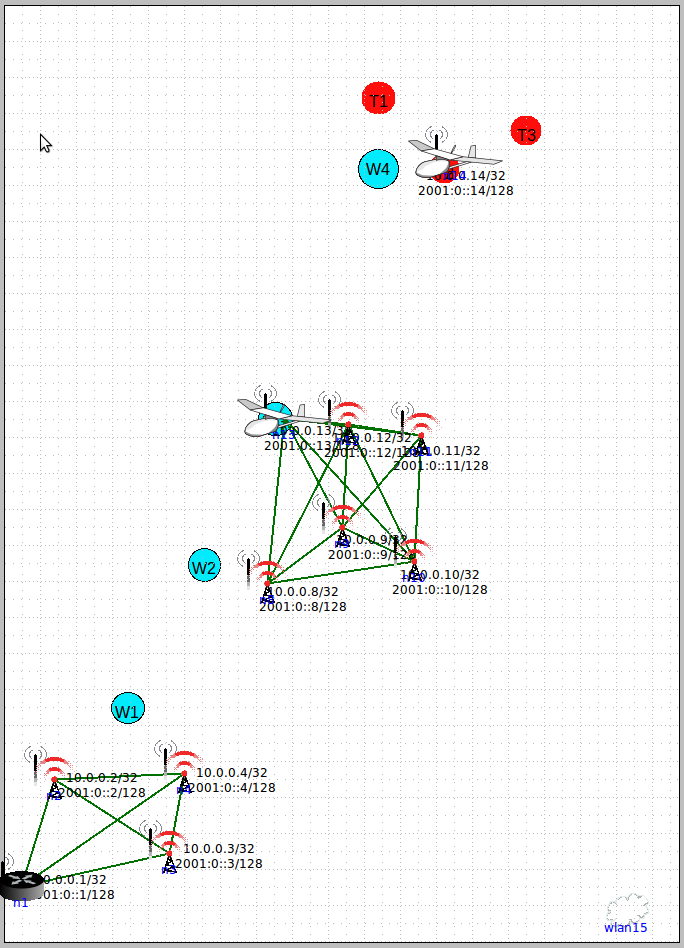}
    \subcaption{Step 6: Nodes 5-7 have failed.}\label{subfig:relays2} 
  \end{subfigure}
  \begin{subfigure}[t]{0.30\linewidth}
    \centering
    \includegraphics[width=\linewidth]{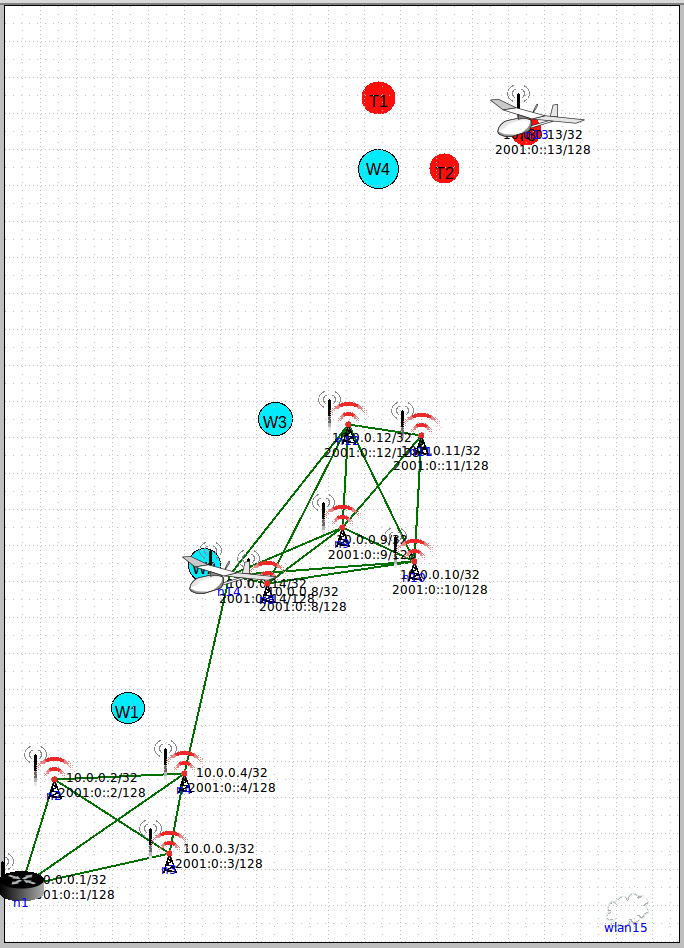}
    \subcaption{Step 7: $u_2$ re-plans, moves closer to home base.}\label{subfig:relays3}
  \end{subfigure}
  \\
  \begin{subfigure}[t]{0.30\linewidth}
    \centering
    \includegraphics[width=\textwidth]{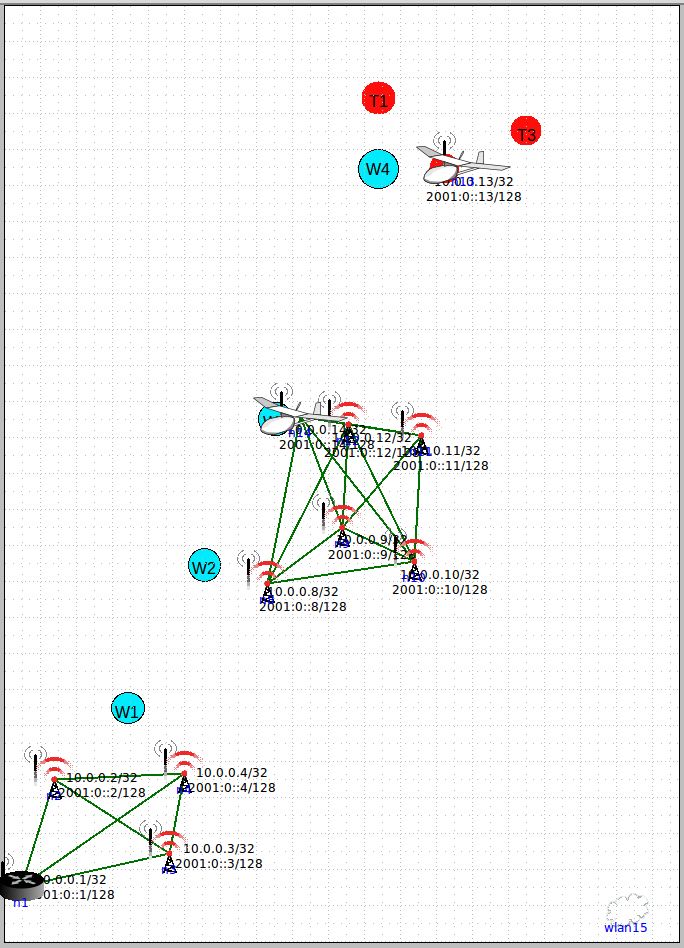}
    \subcaption{Step 8: $u_2$ moves toward $u_1$.}\label{subfig:relays4}
  \end{subfigure}
  \begin{subfigure}[t]{0.30\linewidth}
    \centering
    \includegraphics[width=\textwidth]{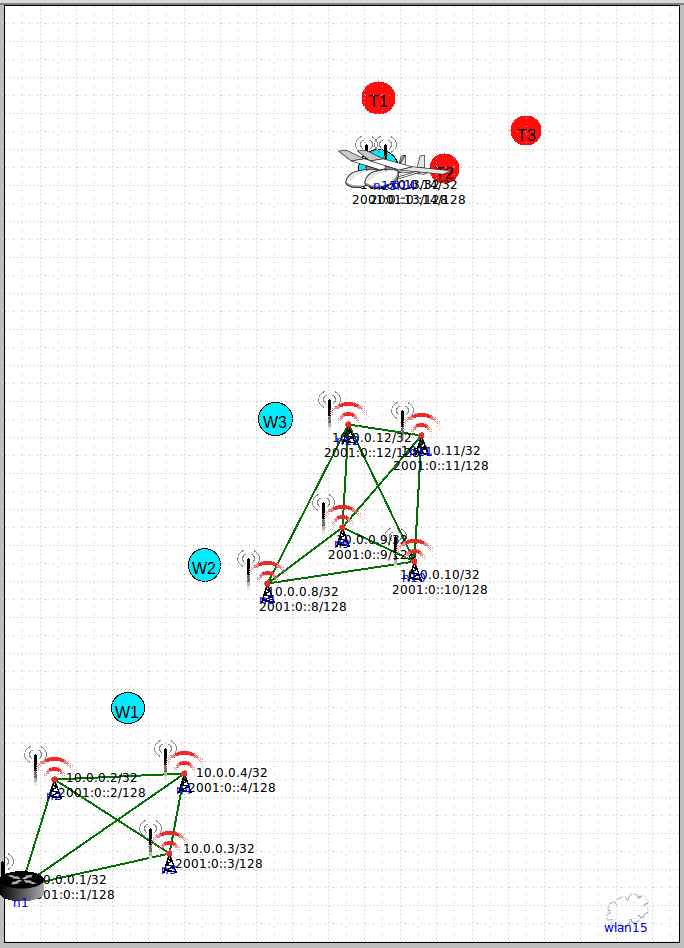}
    \subcaption{Step 9: $u_2$ and $u_1$ reconnect and move back toward home base.}\label{subfig:relays5}
  \end{subfigure}
  \caption{Re-planning after relay node failure between steps 5 and 6 forcing the UAVs to re-plan.}\label{fig:inst3}
\end{figure}
\noindent
\textbf{Example Instance.} Consider the environment shown in in Figure~\ref{fig:inst3}. Two UAVs, $u_1$ and $u_2$ are initially located at the home base in the lower left corner. The home base, relays and targets are positioned as shown in the figure, and the radio range is set to $7$ grid units. 

The mission planner finds a plan in which the UAVs begin by traveling toward the targets. While $u_1$ visits the first two targets, $u_2$ positions itself so as to be in radio contact with $u_1$ (Figure~\ref{subfig:relays}). Upon receipt of the pictures, $u_2$ moves within range of the relays to transmit the pictures to the home base. At the same time, $u_1$ flies toward the final target, where it will be reached by $u_2$ to exchange the final picture.

Now let us consider the impact of unexpected events during mission execution: while $u_2$ is flying back to re-connect with the relays, it observes (``Observe'' step of the architecture from Figure~\ref{fig:arch}) that the home base is unexpectedly not in radio contact (Figure~\ref{subfig:relays2}). Hence, $u_2$ uses the available observations to determine plausible causes (``Explain'' step of the architecture). In this instance, $u_2$ observes that relays $r_5$, $r_6$, $r_7$ and all the network nodes South of them are not reachable via the network. Based on knowledge of the layout of the network, $u_2$ determines that the simplest plausible explanation is that those three relays must have stopped working while $u_2$ was out of radio contact (e.g., started malfunctioning or have been destroyed).
Next, $u_2$ replans (``Local Planner'' step of the architecture). \emph{The plan is created based on the assumption that $u_1$ will continue executing the mission plan. This assumption can be later withdrawn if observations prove it false.} Following the new plan, $u_2$ moves further South towards the home base (Figure~\ref{subfig:relays3}). Simultaneously, $u_1$ continues with the execution of the mission plan, unaware that the connectivity has changed and that $u_2$ has deviated from the mission plan.
After successfully relaying the pictures to the home base, $u_2$ moves back towards $u_1$.  UAV $u_1$, on the other hand, reaches the expected rendezvous point, and observes that $u_2$ is not where expected (Figure~\ref{subfig:relays4}).  UAV $u_1$ does not know the actual position of $u_2$, but its absence is evidence that $u_2$ must have deviated from the  plan at some  point in time. Thus, $u_1$'s must now replan. Not knowing $u_2$'s state, $u_1$'s plan is to fly South to relay the missing picture to the home base on its own. This plan still does not deal with the unavailability of $r_5$, $r_6$, $r_7$, since $u_1$ has not yet had a chance to get in radio contact with the relays and observe the current network connectivity state.
The two UAVs continue with the execution of their new plans and eventually meet, unexpectedly for both (Figure~\ref{subfig:relays5}). At that point, they automatically share the final picture. Both now determine that the mission can be completed by flying South past the failed relays, and execute the corresponding actions.

\section{Conclusion and Future Work} 

This paper discussed a novel application of an ASP-based intelligent agent architecture to the problem of UAV coordination while taking into account network communications. Our work demonstrates the advantages deriving from such network-aware reasoning. In on-going experimental evaluations, our approach yielded a reduction in mission length of up to $30\%$ and in total staleness between $50\%$ and $100\%$. We expect that, in more complex scenarios, the advantage of a realistic networking model will be even more evident.

\bibliographystyle{acmtrans}
\bibliography{maranets-paper,biblio-mb-mod}

\end{document}